\documentclass[10pt,letterpaper]{article}
\usepackage{spconf,amsmath,amssymb,graphicx}
\usepackage[T1]{fontenc}
\usepackage{booktabs,microtype,etoolbox}
\usepackage[hidelinks]{hyperref}
\hypersetup{pdftitle={Video–Time Priors for EEG–fNIRS Emotion Regression on Familiar Videos},pdfauthor={Minghao Kong, Ying Gao, Jiurun Chen, Muyao Chen, Zheng Ge, Jianxin Wu, Rongjie Wang}}

\title{\fontsize{14}{16}\selectfont Video--Time Priors for EEG--fNIRS Emotion Regression on Familiar Videos}
\graphicspath{{figures/}}
\name{\normalsize Minghao Kong$^{1,2}$, Ying Gao$^{2}$, Jiurun Chen$^{2}$, Muyao Chen$^{2}$,
Zheng Ge$^{1,2}$, Jianxin Wu$^{1,*}$, and Rongjie Wang$^{2,*}$\thanks{\textsuperscript{*}Corresponding authors: Jianxin Wu and Rongjie Wang.}\thanks{This work was supported in part by the Major Key Project of Pengcheng
Laboratory under Grant PCL2025A12-3 and by the Science and Technology Planning
Project of Guangdong Science and Technology Department under the Guangdong Key
Laboratory of Advanced IntelliSense Technology under Grant 2019B121203006.}}
\address{\normalsize\begin{tabular}{@{}c@{}}
$^{1}$School of Electronics and Communication Engineering, Sun Yat-sen University, Shenzhen Campus, Shenzhen, China\\
$^{2}$Department of Network Intelligence, Pengcheng Laboratory, Shenzhen, China
\end{tabular}}

\begin{document}
\maketitle

\begin{abstract}
Do viewer-specific physiological signals improve continuous emotion
regression beyond the stimulus-driven response shared across viewers? We
examine this question by predicting emotion ratings for new viewers watching
videos that also appear in the training set. As a stimulus-only baseline, we
define a video--time prior: the smoothed median of training labels for each
video at each second. We then combine this prior, using a fixed fusion weight,
with a graph-based model of electroencephalography (EEG) and functional
near-infrared spectroscopy (fNIRS) signals. On MER-PS 2026, five-fold
cross-validation grouped by participant over 24 participants yields a mean
absolute error (MAE) of 29.01 for fusion, compared with 29.06 for the prior
alone. On an external cohort of four participants excluded from training, the
corresponding MAEs are 27.72 and 28.04. In this cohort, most of the MAE
reduction relative to a global-constant baseline comes from the prior, that
is, from video identity and playback time alone. Fusion lowers MAE for three
of the four held-out participants and for nine of the fifteen videos. The
added value of physiological signals is therefore small and varies across
participants and videos. Because the model uses physiological signals recorded
after each prediction time, it supports only offline prediction.
Generalization to unseen videos remains untested.
\end{abstract}

\begin{keywords}
EEG--fNIRS, emotion regression, video--time prior, multimodal fusion, cross-participant evaluation
\end{keywords}

\section{Introduction}
\label{sec:intro}

Continuous affect estimation recovers changing valence and arousal.
Electroencephalography (EEG) measures electrical activity; functional
near-infrared spectroscopy (fNIRS) captures slower haemodynamic changes.
Their complementarity motivates graph and attention models
\cite{sun2020multimodal,chen2024eegfnirs}, despite cross-participant shifts
\cite{li2018crosssubject}. Yet accuracy need not come primarily from the
new viewer's physiology.

When participants view the same videos, naturalistic stimuli can elicit
synchronized responses \cite{hasson2004intersubject,dmochowski2012correlated}.
Given video identity and playback time, earlier viewers' labels provide a
shared affect trajectory. Subject-held-out evaluation on repeated videos
therefore tests generalization to new viewers of known videos, not to unseen
stimuli. Splits must match this use case \cite{saeb2017usecase}.

We combine a fold-wise video--time prior with EEG--fNIRS through fixed-weight
fusion. A staged comparison separates contributions from global label level,
video identity, within-video time, and physiology. Five-fold evaluation and an
external cohort show that the prior supplies most of the error reduction over
EEG--fNIRS alone.

\section{Method}
\label{sec:method}

\subsection{Task and information boundary}

\begin{figure*}[t]
  \centering
  \includegraphics[width=\textwidth]{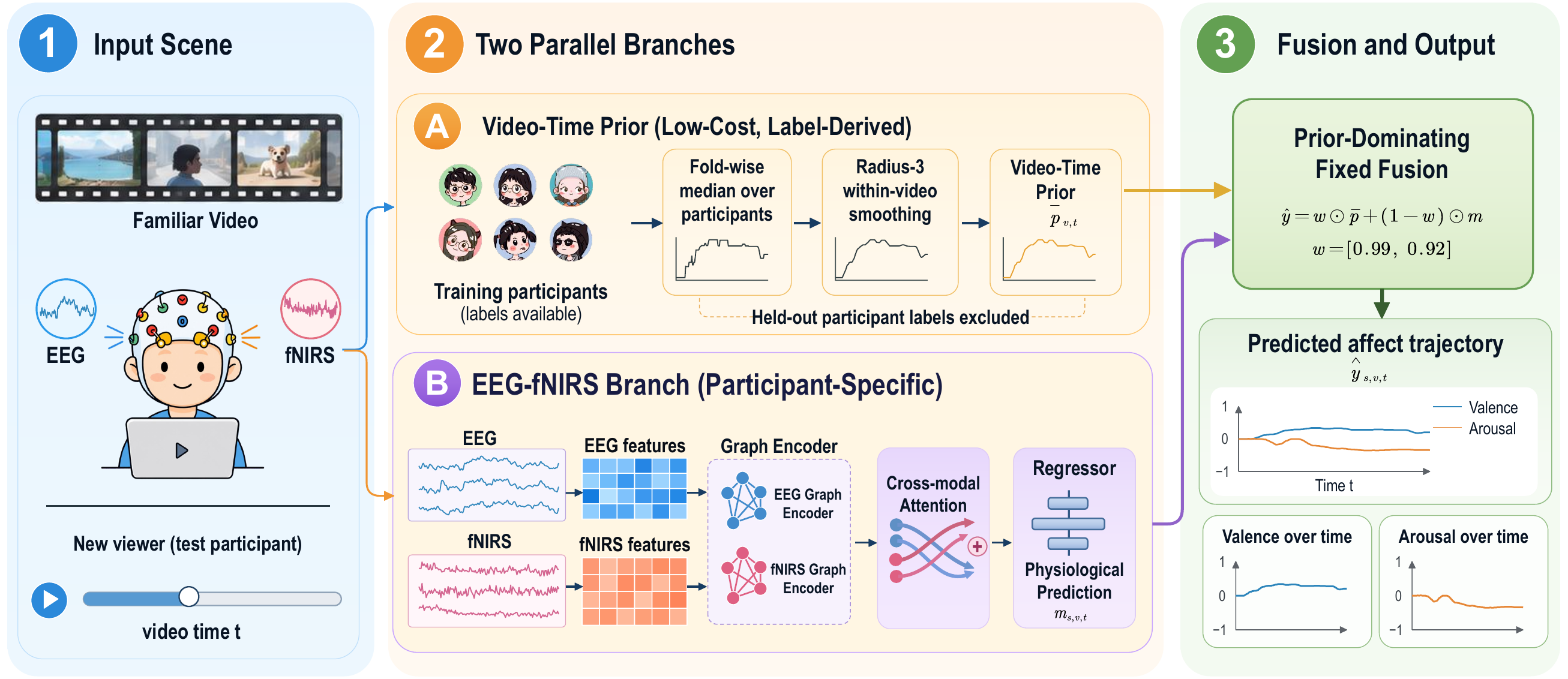}
  \caption{Familiar-video regression. (a) Known video/time and new-viewer
  signals. (b) The fold-wise label prior and offline EEG--fNIRS branch.
  (c) Fixed fusion with prior weights $[0.99,0.92]$ for valence/arousal.
  Held-out labels are excluded; the drawn output curves are schematic.}
  \label{fig:method}
\end{figure*}

For viewer $s$, video $v$ seen during training, and aligned second $t$,
$y_{s,v,t}\in[1,255]^2$ contains valence and arousal. The prior retrieves
a training-label statistic by $(v,t)$; physiology comes from the new viewer.
Held-out labels never enter either branch (Fig.~\ref{fig:method}).

\subsection{Fold-wise video--time prior}

For each target $d\in\{V,A\}$ and training-participant set
$\mathcal S_{\mathrm{tr}}$, the unsmoothed prior is
\begin{equation}
 p^d_{v,t}=\operatorname{median}_{s\in\mathcal S_{\mathrm{tr}}}
 y^d_{s,v,t}.
 \label{eq:median}
\end{equation}
A radius-three moving average smooths each median trajectory:
\begin{equation}
 \begin{split}
 \bar p^d_{v,t}&=\frac{1}{|\mathcal I_{v,t}|}
 \sum_{u\in\mathcal I_{v,t}}p^d_{v,u},\\
 \mathcal I_{v,t}&=\{u:|u-t|\leq3,\ 0\leq u<T_v\}.
 \end{split}
 \label{eq:smooth}
\end{equation}
Here $T_v$ is video length; smoothing remains within video boundaries.

\subsection{Physiological branch}

Signals are centered using the corresponding five-second resting baseline,
and EEG is resampled to 200 Hz. For each channel and one-second EEG segment,
we compute relative log power and differential entropy in six bands spanning
1--45 Hz, plus Hjorth activity, mobility, and complexity
\cite{hjorth1970eeg}. Features from $t-1$, $t$, and $t+1$ are concatenated
into a $64\times45$ tensor. Each fNIRS channel includes six signal types:
HbO, HbR, HbT, and Abs at 780, 805, and 830 nm \cite{merpsdataset}.
For each type, five per-second statistics (mean, standard deviation, slope,
skewness, and kurtosis) are concatenated across the same context into a
$51\times90$ tensor.
Context windows at trial boundaries use edge padding. The $t+1$ segment
makes this an offline estimator.

Each modality uses a learned-adjacency graph encoder with identity,
first-order, and second-order propagation supports and 32-dimensional
channel embeddings. Four-head cross-modal attention operates in both
directions \cite{vaswani2017attention,tsai2019multimodal}. Mean and maximum
pooling over modality-specific and cross-modal tokens feed a multilayer
regressor with a sigmoid output. Targets are scaled as
$\tilde y=(y-1)/254$. Training minimizes mean squared error (MSE) plus
$0.01$ times a symmetric contrastive alignment loss between paired
EEG and fNIRS representations \cite{chen2020simclr}. Dropout is $0.7$;
optimization uses AdamW and early stopping on fold validation MSE.
Feature-standardization statistics come from the corresponding training
participants. Predictions are transformed by $254\hat{\tilde y}+1$ and
clipped to $[1,255]$, yielding $m_{s,v,t}$.

\subsection{Fixed fusion}

We combine the two predictions by
\begin{equation}
 \hat y_{s,v,t}=w\odot\bar p_{v,t}+(1-w)\odot m_{s,v,t},
 \qquad w=[0.99,0.92],
 \label{eq:fusion}
\end{equation}
where $\odot$ is elementwise multiplication and weights are for valence
and arousal. The same weights are applied unchanged to the external cohort.

\begin{figure*}[t]
  \centering
  \makeatletter
  \patchcmd{\@makecaption}{\vskip 10pt}{\vskip 5pt}{}{
    \PackageError{main}{Fig. 2 caption spacing patch failed}{Check the spconf caption definition.}}
  \makeatother
  \includegraphics[width=\textwidth,trim=0 9.5bp 0 7.5bp,clip]{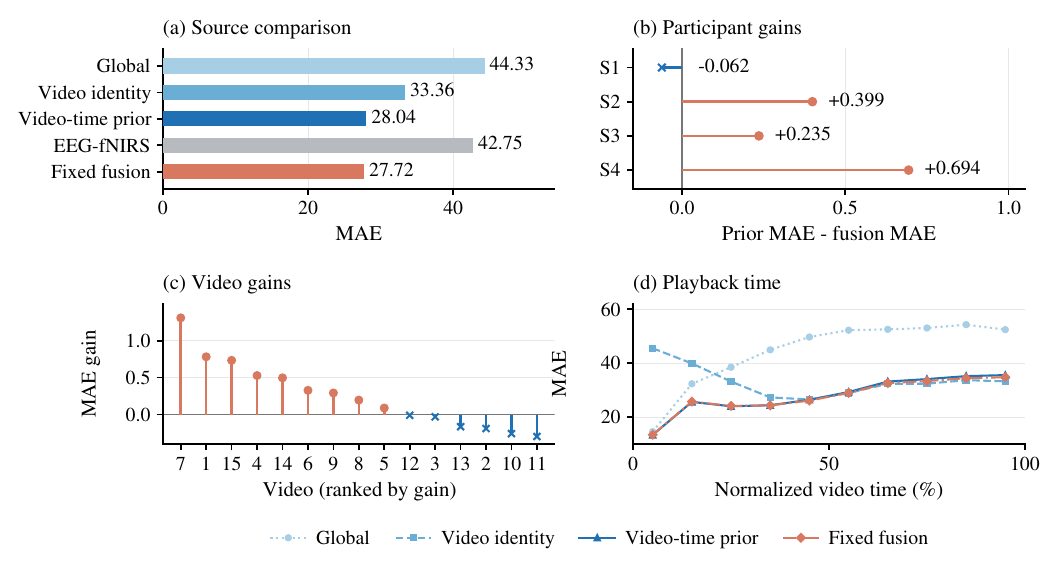}
  \caption{Error decomposition on the external cohort (four participants).
  (a) Overall MAE. (b,c) Paired prior-minus-fusion MAE for all participants
  and videos; positive values favor fusion, with circles marking gains and
  crosses marking losses. (d) Sample-pooled MAE in ten normalized playback
  bins.}
  \label{fig:heterogeneity}
\end{figure*}

\section{Experimental Setup}
\label{sec:setup}

\subsection{Data and evaluation protocols}

MER-PS 2026 development data contain 24 participants viewing 15 videos:
360 trials and 36,864 one-hertz valence--arousal targets
\cite{merpsdataset,lian2026mer}.

\textbf{Internal evaluation.} Participants are partitioned into five folds.
Each prior uses only its fold's training labels, and its matching
EEG--fNIRS checkpoint predicts responses of held-out participants. The
all-development checkpoint (below) is not used. Out-of-fold predictions are pooled before scoring.
The same videos appear in training and held-out folds.

\textbf{External cohort.} Four additional participants supply 60 trials and
6,143 one-hertz targets on the same 15 videos. Their labels are excluded from
training and prediction; predictions are generated before retrospective scoring.
Physiological predictions average outputs from six checkpoints: five fold models
and one model trained on all 24 development participants (the all-development
checkpoint). Each checkpoint uses its own training standardization; the prior
uses all 24 development participants.

\subsection{Source variants and metrics}

Five variants evaluated on the external cohort compare available sources: \emph{global constant}
(two-dimensional median of all development labels), \emph{video identity}
(temporal median of each video's smoothed prior), \emph{video--time prior}
(Eqs.~\ref{eq:median}--\ref{eq:smooth}), \emph{EEG--fNIRS} (physiology without
explicit video/time lookup), and \emph{fixed fusion} (Eq.~\ref{eq:fusion}).
Internal evaluation reports only the latter three.

\begingroup
\setlength{\abovedisplayskip}{4pt}
\setlength{\belowdisplayskip}{4pt}
\setlength{\abovedisplayshortskip}{4pt}
\setlength{\belowdisplayshortskip}{4pt}
The primary endpoint is sample-pooled MAE on the original label scale:
\begin{equation}
 \operatorname{MAE}=\frac{1}{2N}\sum_{i=1}^{N}
 \sum_{d\in\{V,A\}}|y_i^d-\hat y_i^d|,
 \label{eq:mae}
\end{equation}
where $i$ indexes participant--video--second tuples.
Internal scores use floating-point predictions. To match
our released inference code, predictions from each external variant are rounded
to the nearest integer and clipped to $[1,255]$ before scoring.
Paired prior and fusion errors are aggregated by participant, video, and ten
equal bins of normalized playback time. These summaries are descriptive:
the external cohort has only four participant clusters, and correlated seconds
are not treated as independent replicates for significance tests.
\par\endgroup

\begin{figure*}[t]
  \centering
  \includegraphics[width=\textwidth]{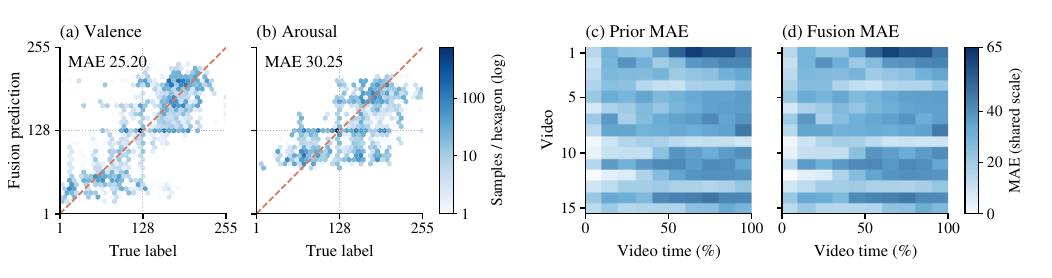}
  \caption{Prediction and error diagnostics on the external cohort using all 6,143
  one-second samples. (a,b) Fixed-fusion prediction versus target, with a
  shared log-count density scale and ideal diagonal. (c,d) Prior and fusion
  MAE over all 15 videos and ten time bins, using the same $0$--$65$ scale.}
  \label{fig:diagnostics}
\end{figure*}

\section{Results and Discussion}
\label{sec:results}

\begin{table}[t]
  \centering
  \caption{MAE on the original $[1,255]$ scale (lower is better).
  Bold marks the lowest MAE.}
  \label{tab:main}
  \small
  \setlength{\tabcolsep}{5pt}
  \begin{tabular}{@{}lrrr@{}}
    \toprule
    Variant & Overall & Valence & Arousal \\
    \midrule
    \multicolumn{4}{@{}l}{\textit{Internal: five folds, 24 participants}}\\
EEG--fNIRS & 47.35 & 51.35 & 43.35 \\
Video--time prior & 29.06 & 26.67 & 31.46 \\
Fixed fusion & \textbf{29.01} & \textbf{26.66} & \textbf{31.37} \\
\midrule
\multicolumn{4}{@{}l}{\textit{External cohort: 4 participants}}\\
Global constant & 44.33 & 47.46 & 41.20 \\
Video identity & 33.36 & 32.56 & 34.17 \\
EEG--fNIRS & 42.75 & 45.29 & 40.21 \\
Video--time prior & 28.04 & 25.34 & 30.74 \\
Fixed fusion & \textbf{27.72} & \textbf{25.20} & \textbf{30.25} \\
\bottomrule
  \end{tabular}
\end{table}

\subsection{Prediction accuracy and source contributions}

Fixed fusion has the lowest observed MAE in both evaluations
(Table~\ref{tab:main}). It improves on the prior by only 0.05 internally
and 0.32 externally, with the same ordering for valence and arousal.

The staged comparison decomposes the observed MAE reductions. On the external cohort,
replacing the global constant with video identity lowers overall MAE by
10.97 points; adding within-video time lowers it by another 5.32.
Fusion then lowers MAE by only 0.32 points. Using MAEs before display rounding,
the prior accounts for 99.73\% and 97.89\% of the reductions from EEG--fNIRS
to fusion internally and on the external cohort, respectively. These ratios
describe error, not information content.

\subsection{Heterogeneity of physiological correction}

Define fusion gain as
$\Delta=\mathrm{MAE}_{\mathrm{prior}}-\mathrm{MAE}_{\mathrm{fusion}}$.
Positive values favor fusion. Fig.~\ref{fig:heterogeneity}(b) shows lower MAE
for three of four participants in the external cohort. Fusion lowers MAE for
nine of fifteen videos; the largest gain is $+1.31$ for V7, and the largest
loss is $-0.30$ for V11. Across participant--video trials, fusion lowers MAE
in 38 and raises it in 22. It raises MAE in the first three normalized time
bins but lowers it in the remaining seven.

The benefit of time conditioning is also nonuniform. Video identity has
45.53 MAE versus 13.40 for the prior in the first bin, but lower MAE in each
of the last five. The prior's 11.89-point pooled MAE reduction in the first
five bins outweighs its 1.33-point increase in the last five, yielding the
5.32-point overall reduction.

Fig.~\ref{fig:diagnostics}(a,b) reveals compressed prediction ranges and
deviations from the ideal diagonal that aggregate MAE does not show.

The video-by-time heatmaps in Fig.~\ref{fig:diagnostics}(c,d) show
shared error structure. Across the
150 cells formed by 15 videos and ten normalized bins, fusion has lower
MAE in 76, equal MAE in seven, and higher MAE in 67. The methods share
nine of their ten highest-error cells, with a common hotspot in the latter
half of V1. Much of the prior's error structure persists after fusion.

\subsection{Weight sensitivity and limits of interpretation}

A post hoc sweep of prior weights over $[0,1]$ in $0.01$ steps gives internal
target-wise minima at $(0.99,0.92)$. At these weights, fusion lowers MAE
relative to the prior in only three of five folds. This sweep does not
reproduce the original selection procedure. Minima on the external cohort
shift to $(0.83,0.82)$ (MAE 26.91); these are diagnostic, and the reported
fusion results retain the original fixed weights.

These results concern videos seen during training; unseen-video and cross-site
transfer were not evaluated. Checkpoint and weight selection was not fully
nested; future work should nest both within outer participant splits
\cite{varma2006bias}. Internal and external gains are not directly comparable
because the ensembles (one versus six checkpoints per prediction), rounding,
and held-out groups differ.

Only one physiological architecture is evaluated, without isolating EEG,
fNIRS, graph, or attention effects. The result does not establish or rule out
independent affective information in either modality. No temporal lag
compensates for the fNIRS haemodynamic delay. Causal features and crossed
participant-by-video holdouts are needed to extend the findings.

The prior requires earlier labels but avoids new physiological acquisition;
latency and energy savings are not measured. Future work should model the
residual and test when individual corrections help.

\section{Conclusion}

A fold-wise video--time prior supplies most of fixed fusion's advantage
over EEG--fNIRS alone. The physiological correction is small and
heterogeneous. Familiar-video evaluation should therefore separate shared
stimulus information from individual correction.

\noindent\textbf{Reproducibility.} The implementation and evaluation scripts
are available under Apache 2.0 at
\url{https://github.com/rudykon/Vtp-Emotion}.
Dataset access follows the corresponding MER-PS repository terms.

\clearpage
\patchcmd{\thebibliography}{\section{References}}
  {\section*{\centering\normalsize REFERENCES}}{}{
    \PackageError{main}{Unnumbered reference heading patch failed}{Check the bibliography definition.}}


\begin{thebibliography}{10}

\bibitem{sun2020multimodal}
Yanjia Sun, Hasan Ayaz, and Ali~N. Akansu,
\newblock ``Multimodal affective state assessment using f{NIRS} + {EEG} and
  spontaneous facial expression,''
\newblock {\em Brain Sciences}, vol. 10, no. 2, pp. 85, 2020.

\bibitem{chen2024eegfnirs}
Guijun Chen, Yue Liu, and Xueying Zhang,
\newblock ``{{EEG}--f{NIRS}}-based emotion recognition using graph convolution
  and capsule attention network,''
\newblock {\em Brain Sciences}, vol. 14, no. 8, pp. 820, 2024.

\bibitem{li2018crosssubject}
Xiang Li, Dawei Song, Peng Zhang, Yazhou Zhang, Yuexian Hou, and Bin Hu,
\newblock ``Exploring {EEG} features in cross-subject emotion recognition,''
\newblock {\em Frontiers in Neuroscience}, vol. 12, pp. 162, 2018.

\bibitem{hasson2004intersubject}
Uri Hasson, Yuval Nir, Ifat Levy, Galit Fuhrmann, and Rafael Malach,
\newblock ``Intersubject synchronization of cortical activity during natural
  vision,''
\newblock {\em Science}, vol. 303, no. 5664, pp. 1634--1640, 2004.

\bibitem{dmochowski2012correlated}
Jacek~P. Dmochowski, Paul Sajda, Joao Dias, and Lucas~C. Parra,
\newblock ``Correlated components of ongoing {EEG} point to emotionally laden
  attention---a possible marker of engagement?,''
\newblock {\em Frontiers in Human Neuroscience}, vol. 6, pp. 112, 2012.

\bibitem{saeb2017usecase}
Sohrab Saeb, Luca Lonini, Arun Jayaraman, David~C. Mohr, and Konrad~P. Kording,
\newblock ``The need to approximate the use-case in clinical machine
  learning,''
\newblock {\em GigaScience}, vol. 6, no. 5, pp. 1--9, 2017.

\bibitem{hjorth1970eeg}
Bo~Hjorth,
\newblock ``{EEG} analysis based on time domain properties,''
\newblock {\em Electroencephalography and Clinical Neurophysiology}, vol. 29,
  no. 3, pp. 306--310, 1970.

\bibitem{merpsdataset}
{{MER-PS}},
\newblock ``{MER-PS/MER-PS-trainval},'' Hugging Face Datasets. [Online].
  Available: \url{https://huggingface.co/datasets/MER-PS/MER-PS-trainval},
  2026,
\newblock Dataset repository, accessed 23 July 2026.

\bibitem{vaswani2017attention}
Ashish Vaswani, Noam Shazeer, Niki Parmar, Jakob Uszkoreit, Llion Jones,
  Aidan~N. Gomez, {\L}ukasz Kaiser, and Illia Polosukhin,
\newblock ``Attention is all you need,''
\newblock in {\em Advances in Neural Information Processing Systems}, Red Hook,
  NY, USA, 2017, vol.~30, pp. 5998--6008, Curran Associates, Inc.

\bibitem{tsai2019multimodal}
Yao-Hung~Hubert Tsai, Shaojie Bai, Paul~Pu Liang, J.~Zico Kolter,
  Louis-Philippe Morency, and Ruslan Salakhutdinov,
\newblock ``Multimodal transformer for unaligned multimodal language
  sequences,''
\newblock in {\em Proceedings of the 57th Annual Meeting of the Association for
  Computational Linguistics}, Florence, Italy, 2019, pp. 6558--6569,
  Association for Computational Linguistics.

\bibitem{chen2020simclr}
Ting Chen, Simon Kornblith, Mohammad Norouzi, and Geoffrey Hinton,
\newblock ``A simple framework for contrastive learning of visual
  representations,''
\newblock in {\em Proceedings of the 37th International Conference on Machine
  Learning}, Virtual, 2020, vol. 119 of {\em Proceedings of Machine Learning
  Research}, pp. 1597--1607, PMLR.

\bibitem{lian2026mer}
Zheng Lian, Xiaojiang Peng, Kele Xu, Ziyu Jia, Xinyi Che, Zebang Cheng, Fei Ma,
  Laizhong Cui, Yazhou Zhang, Xin Liu, Liang Yang, Jia Li, Fan Zhang, Liumeng
  Xue, Erik Cambria, Guoying Zhao, Bjorn~W. Schuller, and Jianhua Tao,
\newblock ``{MER} 2026: From discriminative emotion recognition to generative
  emotion understanding,'' arXiv:2604.19417, 2026.

\bibitem{varma2006bias}
Sudhir Varma and Richard Simon,
\newblock ``Bias in error estimation when using cross-validation for model
  selection,''
\newblock {\em BMC Bioinformatics}, vol. 7, no. 1, pp. 91, 2006.

\end{thebibliography}
\end{document}